\pdfoutput=1

\documentclass[11pt]{article}

\usepackage[final]{acl}

\usepackage{times}
\usepackage{latexsym}

\usepackage[T1]{fontenc}

\usepackage[utf8]{inputenc}

\usepackage{microtype}

\usepackage{inconsolata}

\usepackage{graphicx}
\usepackage{amsmath}
\usepackage{booktabs}
\usepackage{multirow}

\title{Learning to Align Multi-Faceted Evaluation:\\ A Unified and Robust Framework}

\author{Kaishuai Xu$^{1}$\thanks{This work was done during an internship at Huawei Noah’s Ark Lab.} , Tiezheng Yu$^{2}$, Wenjun Hou$^{1}$, Yi Cheng$^{1}$, \\
\textbf{Liangyou Li$^{2}$, Xin Jiang$^{2}$, Lifeng Shang$^{2}$, Qun Liu$^{2}$, Wenjie Li$^{1}$}  \\
$^1$The Hong Kong Polytechnic University
$^2$Huawei Noah’s Ark Lab
\\
\texttt{kaishuaii.xu@connect.polyu.hk} 
}

\begin{document}
\maketitle

\begin{abstract}

Large Language Models (LLMs) are being used more and more extensively for automated evaluation in various scenarios. 
Previous studies have attempted to fine-tune open-source LLMs to replicate the evaluation explanations and judgments of powerful proprietary models, such as GPT-4. However, these methods are largely limited to text-based analyses under predefined general criteria, resulting in reduced adaptability for unseen instructions and demonstrating instability in evaluating adherence to quantitative and structural constraints. 
To address these limitations, we propose a novel evaluation framework, ARJudge, that adaptively formulates evaluation criteria and synthesizes both text-based and code-driven analyses to evaluate LLM responses. ARJudge consists of two components: a fine-tuned Analyzer that generates multi-faceted evaluation analyses and a tuning-free Refiner that combines and refines all analyses to make the final judgment. We construct a Composite Analysis Corpus that integrates tasks for evaluation criteria generation alongside text-based and code-driven analysis generation to train the Analyzer. 
Our results demonstrate that ARJudge outperforms existing fine-tuned evaluators in effectiveness and robustness. Furthermore, it demonstrates the importance of multi-faceted evaluation and code-driven analyses in enhancing evaluation capabilities.

\end{abstract}
\section{Introduction}
\begin{figure}[t!]
	\centering
	\includegraphics[width=1.0\linewidth]{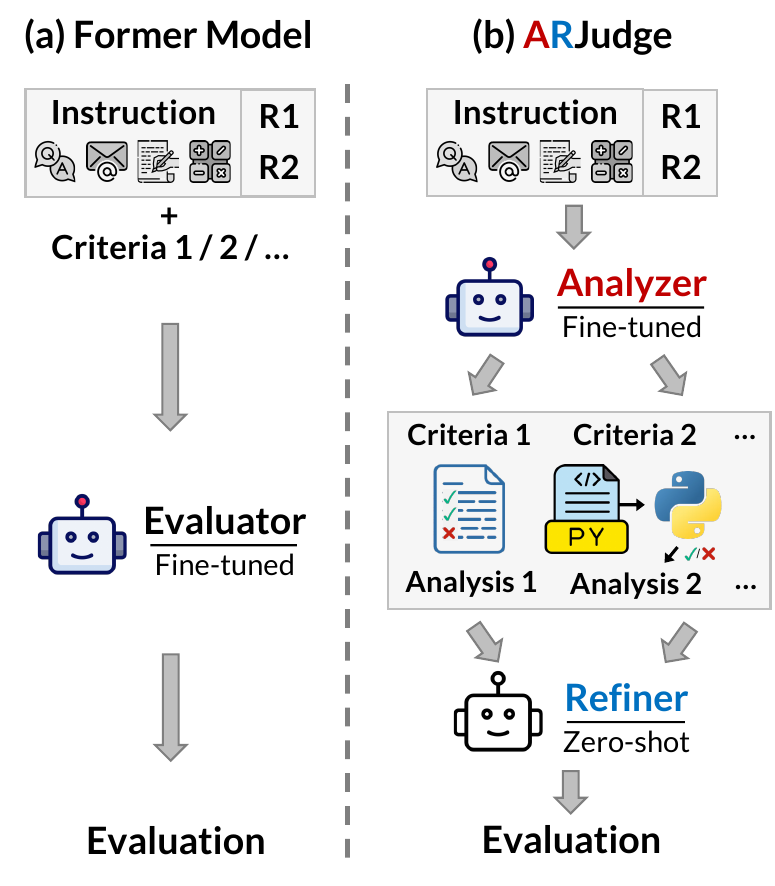}
	\caption{Comparison of previous fine-tuned evaluators and our framework. \textbf{Left} is a former model and \textbf{Right} is our ARJudge. The Analyzer adaptively defines evaluation criteria and conducts multi-faceted analyses in various forms, e.g., text or code. The Refiner combines all preceding analyses and produces the final evaluation.}
	\label{intro_example}
\end{figure}

The rapid advancement of Large Language Models (LLMs) has highlighted the critical need for robust output evaluation methods \cite{survey_judge}. While proprietary models like GPT-4 have emerged as predominant evaluation approaches given their superior capabilities, transparent and controllable considerations have driven research toward fine-tuning open-source LLMs for evaluation tasks \cite{prometheus, prometheus2}. Recent work has established the viability of open-source alternatives by training LLMs to replicate the evaluation explanations and judgments of proprietary models \cite{critiqueLLM, x-eval, themis, prometheus2}.

However, existing fine-tuned evaluators rely solely on text-based analysis with predefined evaluation criteria, leading to two key limitations \cite{auto-j, themis, judgelm, prometheus2}. 
First, evaluation based on predefined criteria can not fully capture the nuanced task requirements. 
For example, general criteria for writing, such as conciseness or logical structure, may not be sufficient for evaluating creative writing tasks that require an engaging plot. Moreover, it is challenging to effectively adapt predefined criteria to new and diverse instructions \cite{auto-j}. 
Second, LLM-based evaluators demonstrate significant instability in evaluating adherence to complex instruction requirements, particularly objective criteria such as quantitative or structural constraints \cite{ifeval}. For instance,  they struggle to reliably assess basic textual attributes such as word counts, a common requirement in writing-related instructions \cite{llm_bad_word_level}. These limitations also extend to the evaluation of formatting constraints. 

In this work, we argue that developing robust fine-tuned evaluators requires the ability to adaptively generate evaluation criteria and conduct multi-faceted analyses \cite{branch-merge}. These abilities enhance the evaluators' comprehensive performance in both what to evaluate and how to evaluate. Even for unseen instructions, the evaluators can define tailored criteria and assess instructions with nuanced precision.   
Furthermore, evaluators should use automated tools to assess objective requirements \cite{mint}. These tools provide reproducible feedback, offering reliable verification that helps overcome LLMs' inherent limitations in objective evaluation. 

To address these challenges, we propose ARJudge, a novel evaluation framework that combines adaptive criteria generation with text-based and code-driven analysis generation to comprehensively assess LLM outputs. ARJudge comprises two core components: (1) an Analyzer that generates multi-faceted evaluation with text-based and code-driven analyses and (2) a Refiner that synthesizes and refines these analyses to produce well-reasoned judgments. 
We train ARJudge on a curated Composite Analysis Corpus, which contains tasks for generating evaluation criteria and performing multi-faceted analyses in both text and code. This corpus enables the Analyzer to learn context-sensitive evaluation logic, such as deriving criteria from instructions and assessing responses accordingly. Extensive experiments across multiple benchmarks demonstrate ARJudge’s superiority and robustness over existing open-source evaluators. Our further analysis validates the necessity and effectiveness of integrating code-driven analyses, which improve accuracy in evaluating instruction following by up to 11.1\% compared to text-only methods. 

The main contributions of this work include:
\begin{itemize}
	\item We propose ARJudge, a novel evaluation framework that combines adaptive criteria generation with text-based and code-driven analyses to evaluate LLM outputs. By incorporating code-driven analytical capabilities, ARJudge extends beyond traditional text-based evaluation approaches.
	\item We develop a training dataset, Composite Analysis Corpus, containing samples for evaluation criteria generation, text-based analyses, and code-driven analyses. It is the first dataset to incorporate multi-faceted analytical samples for evaluator training.
	\item Extensive experiments across multiple benchmarks demonstrate ARJudge’s superior performance over existing fine-tuned evaluators.
\end{itemize}

\section{Composite Analysis Corpus}

\begin{figure*}[t!]
	\centering
	\includegraphics[width=0.9\textwidth]{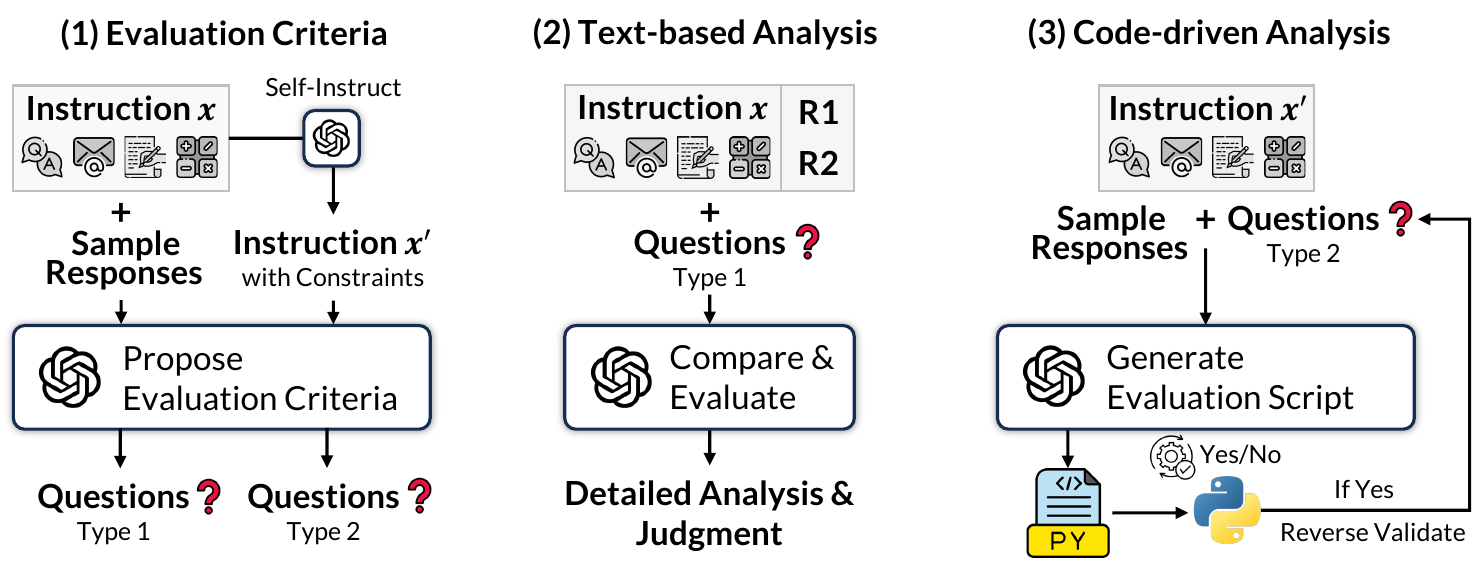}
	\caption{The overview of the corpus construction. ``\textbf{R1}'' and ``\textbf{R2}'' denote two candidate responses with a preference annotation. ``\textbf{Sample Responses}'' are newly sampled responses that we use as references to generate evaluation questions and code scripts. Step (1) produces two types of evaluation questions, respectively. Step (2) and Step (3) develop corresponding text-based and code-driven analyses.}
	\label{corpus}
\end{figure*}

Collecting comprehensive and detailed evaluation analysis data is essential for fine-tuning an LLM to improve evaluation performance \cite{auto-j, themis}. Previous studies \cite{auto-j, themis, prometheus, prometheus2} focus exclusively on text-based analysis with predefined general evaluation criteria, showing limited generalization and robustness \cite{sft_eval_limit}. To address these limitations, we develop a composite analysis corpus to improve LLMs' ability to determine what to evaluate and how to evaluate effectively. The process of constructing the corpus involves three steps: (1) establishing evaluation criteria specifically for each instruction (§\ref{subsec:criteria}), (2) conducting text-based analyses to assess responses using multiple criteria (§\ref{subsec:text}), and (3) designing code-driven analyses to assess whether responses satisfy the objective requirements of the instructions (§\ref{subsec:code}). 

First of all, we collect a large set of instructions from publicly available preference datasets based on \citet{auto-j}. These datasets \cite{mtbench, webgpt, gptj-pair, saferlhf} consist of preference pairs of LLM-generated responses to identical instructions. Each pair is annotated with a preference label that identifies the better response. In line with \citet{auto-j}, non-English instructions and multi-turn interactions are removed. Then, we establish multiple evaluation criteria for each instruction.

\subsection{Establishing Evaluation Criteria}\label{subsec:criteria}
We define the evaluation criteria in the form of concise questions \cite{llmbar, prometheus2}. Each question describes one aspect that a high-quality response should fulfill. For example, responses to the instruction ``Draft an email to my deputy chairperson requesting their signature on the attached approval letter in a professional and polite manner'' can be evaluated using the following three questions: `\textit{`1. Does the response include a polite and professional request for the deputy chairperson to sign the attached approval letter? 2. Does the response mention the attached approval letter and provide the necessary details about it? 3. Does the response offer assistance with any questions or clarifications the deputy chairperson might have about the approval letter?}'' We establish two types of questions by prompting an LLM in a zero-shot manner. \textbf{Type 1} focuses on generating text-based analysis, while \textbf{Type 2} involves generating Python functions and using execution feedback as code-driven analysis.

To generate the first type of question, we prompt an LLM using three sample responses produced by advanced LLMs as well as the instruction $x$. Such sample responses offer a reference understanding of the instruction. The specific prompt is shown in Figure \ref{prompt_question}. We collect three questions $q_{text}$ for each instruction $x$ following \citet{llmbar} and construct training samples in the format $(x, q_{text})$.

For the second type, we must generate new instructions $x'$ with objective constraints in advance, since their proportion in the datasets is relatively low. We use the self-instruct \cite{self-instruct} method to add objective constraints to the instructions and then produce evaluation questions for verifying these constraints. Following the verifiable instructions\footnote{Verifiable instructions are instructions that can be objectively verified for compliance using tools.} summarized by \citet{ifeval}, we first generate several objective constraints for each instruction, such as ``word count'' and ``end with''. The specific prompt is shown in Figure \ref{prompt_constraint}. Then, we randomly select one to three constraints to add to each instruction and collect the corresponding evaluation questions $q_{code}$. The training samples are constructed in the format $(x', q_{code})$.

\subsection{Collecting Text-based Analysis}\label{subsec:text}
We perform pairwise text-based analyses by providing an LLM with the instruction $x$, two responses $r_1$ and $r_2$, and their corresponding evaluation questions $\{ q_{text} \}$. The output necessitates a comparative analysis for each question, followed by a final determination of the better response. The specific prompt is shown in Figure \ref{prompt_text_eval}. We exclude analyses where the final decision contradicts existing human annotations in the datasets. The training samples are constructed in the format $(x \oplus r_1 \oplus r_2 \oplus q_{text}, y_{text})$. Here, $y_{text}$ denotes the associated analysis result for the evaluation question $q_{text}$, which begins with the hint: ``\textit{Let's evaluate whether responses meet the criteria}''.

\subsection{Developing Code-driven Analysis}\label{subsec:code}
To enhance evaluation robustness, we develop code-driven analyses to assess evaluation questions designed to verify objective requirements. The process is completed in two steps: \textbf{Collecting Python Scripts} and \textbf{Reverse Validation}. The first step involves generating Python functions to analyze whether a response satisfies the objective requirement included in an evaluation question. The second step reversely checks whether the generated function's code is designed to analyze the evaluation question. 

\paragraph{Collecting Python Scripts.}
Given three sample responses and one evaluation question $q_{code}$, we prompt an LLM to generate a Python function for verifying the compliance of sample responses. The input of the function is one response, and the output is a comprehensive intermediate of the results related to the evaluation questions. 
To ensure good generalization, the sample responses are a mix of outputs from advanced and weak LLMs. The specific prompt is shown in Figure \ref{prompt_code_eval}. After prompting, we extract the generated Python function using Markdown parsing. We preliminarily filter out invalid code using two checks: 1. The written Python function fails to execute with the provided sample responses as input; 2. The function fails when tested with an additional set of three sample responses. By filtering out invalid code, we ensure that the generated Python functions are executable and generalizable. 

\paragraph{Reverse Validation.}
To further validate whether the generated Python functions fulfill their intended purpose, we design a reverse validation process. Specifically, we first prompt an LLM with the plain text of the evaluation function, requesting an explanation of the expected outputs. Second, we prompt the LLM again to check for consistency between the explanation and its associated question:
\begin{equation}
\begin{aligned}
e &\sim LLM(f, \text{prompt}_{explain}) \\
r &= LLM(e, q_{code}, \text{prompt}_{check})
\end{aligned}
\end{equation}
where $f$ is the evaluation function, $e$ denotes the generated explanation, $r$ indicates whether the explanation is consistent with the question $q_{code}$. The specific prompts are included in Figure \ref{prompt_reverse}. If the function is found to be inconsistent with the aim of the evaluation question, it is discarded. Finally, we collect the effective Python functions and construct training samples in the format $(x' \oplus r_1 \oplus r_2 \oplus q_{code}, y_{code})$. Here, $y_{code}$ represents the Python function $f$ concatenated with the code output hint ``\textit{Let's write a Python function}''. 

\section{ARJudge}
After constructing the corpus, we collect around 25K composite training samples. We fine-tune an LLM based on them and develop \textbf{ARJudge}, a novel evaluation framework that adaptively evaluates LLM-generated responses and integrates both text-based and code-driven analyses. ARJudge consists of two components: a fine-tuned \textbf{Analyzer} and a tuning-free \textbf{Refiner}. Figure \ref{intro_example} presents the overall framework. The Analyzer is trained on the Composite Analysis Corpus to adaptively generate evaluation criteria for any instruction and produce multi-faceted evaluation, including both text-based and code-driven analyses. The Refiner leverages the general LLM's generalist evaluation capabilities to refine the analysis results produced by the Analyzer and make the final judgment. This framework partially preserves the generalist evaluation pattern of the general model while enhancing the evaluation pattern in the fine-tuning dataset.

\subsection{Training}
We train the Analyzer with diverse training samples and tasks, including question generation samples $(x, q_{text})$ and $(x', q_{code})$, text-based analysis samples $(x \oplus r_1 \oplus r_2 \oplus q_{text}, y_{text})$, and code-driven analysis samples $(x' \oplus r_1 \oplus r_2 \oplus q_{code}, y_{code})$. By training on these combined samples, we aim to enhance the LLM's comprehensive analytical capabilities, enabling it to adaptively propose evaluation criteria and conduct multi-faceted analyses. We employ distinct prompt templates for question generation and response analyses, while maintaining a consistent prompt template for both text-based and code-driven analyses. Different forms of analyses are triggered by their respective starting hints. 

\subsection{Evaluation}
Given an instruction $x$ and two responses $r_1$ and $r_2$, the Analyzer first generates several evaluation questions. Then, it performs a comparative analysis of the two responses based on each evaluation question. Notably, the Analyzer autonomously determines whether to generate Python functions according to question characteristics. If the analysis text includes Python functions, the Analyzer will call a Python interpreter to execute them and return the execution feedback as the code-driven analysis results. Finally, the above multi-faceted analysis results are aggregated and sent to the Refiner for further evaluation. We instruct the Refiner to evaluate the above analysis and refine it with a renewed focus on the instruction's requirements. The Refiner will determine which response is better in a zero-shot manner. 
\section{Experiments}

\begin{table*}[ht!]
\center
\resizebox{0.8\textwidth}{!}{
\begin{tabular}{@{}lccccccccccc@{}}
\toprule
\multirow{3}{*}{Models} & \multicolumn{2}{c}{JudgeLM Eval} & \multicolumn{2}{c}{PandaLM Eval} & \multicolumn{2}{c}{Auto-J Eval} & \multicolumn{2}{c}{MTBench} & \multicolumn{2}{c}{LLMBar} & \multirow{3}{*}{Ave} \\ \cmidrule{2-11} 
                        & Acc             & Agr             & Acc             & Agr            & Acc            & Agr            & Acc          & Agr          & Acc                & Agr  &             \\ \midrule
\textit{Tuning-free}                                                                                                                                                                                      \\ \midrule
GPT-4o                  & 81.8            & 88.1            & 83.1            & 87.5           & 78.6           & 82.5           & 78.8         & 85.4         & 79.8               & 83.4 & 80.4        \\
Claude-3.5-Sonnet       & 82.9            & 86.4            & 86.4            & 91.4           & 78.2           & 85.5           & \textbf{80.8}& 89.1         & \textbf{83.4}      & 90.3 & \textbf{82.3} \\
Deepseek-v3             & \textbf{83.2}   & 85.9            & \textbf{87.4}   & 87.8           & \textbf{82.9}  & 84.2           & 79.7         & 87.0         & 68.6               & 81.6 & 80.4        \\
Qwen2.5-7B              & 80.0            & 78.0            & 80.7            & 79.2           & 73.8           & 65.1           & 75.2         & 72.1         & 52.6               & 65.7 & 72.5        \\ \midrule
\textit{Fine-tuned}                                                                                                                                                                                     \\ \midrule
PandaLM-7B              & 69.9            & 74.7            & 73.1            & 77.8           & 65.2           & 71.0           & 74.0         & 78.4         & 25.9               & 82.5 & 61.6        \\
Auto-J-13B              & 77.9            & 86.6            & 77.2            & 87.2           & \textbf{79.7}  & 87.5           & 75.0         & 84.2         & 27.8               & 83.6 & 67.5        \\
Prometheus2-7B          & 76.5            & 80.3            & 76.3            & 70.9           & 75.1           & 77.2           & 74.3         & 79.5         & 41.5               & 77.6 & 68.7        \\
JudgeLM-7B              & \textbf{81.8}   & 86.0            & 70.3            & 81.4           & 66.1           & 80.2           & 64.6         & 77.1         & 28.1               & 82.0 & 62.2        \\
Themis-8B               & 66.4            & -               & 61.3            & -              & 39.2           & -              & 34.9         & -            & 26.6               & -    & 45.7        \\
ARJudge                 & 81.0            & 83.3            & \textbf{82.4}   & 83.5           & 78.5           & 80.3           & \textbf{78.3}& 81.3         & \textbf{68.2}      & 72.9 & \textbf{77.7}  \\ \bottomrule
\end{tabular}
    }
\caption{Results of different evaluators on the pairwise comparison. ``\textbf{Acc}'' and ``\textbf{Agr}'' denote average accuracy and positional agreement rate. ``\textbf{Ave}'' is the average ``Acc'' across all test sets. The highest average accuracy is marked by \textbf{bold} for two series models, respectively.}
\label{main_result}
\end{table*}

\begin{table}[ht!]
\center
\resizebox{1.0\linewidth}{!}{
\begin{tabular}{@{}lcccc@{}}
\toprule
\multirow{3}{*}{Models} & \multicolumn{4}{c}{LLMBar}           \\ \cmidrule{2-5} 
                        & Neighbor & GPTInst & GPTOut & Manual \\ \midrule
\textit{Tuning-free}                                             \\ \midrule
GPT-4o                  & 81.0     & 86.4    & 75.5   & 76.1   \\
Claude-3.5-Sonnet       & \textbf{83.2}     & \textbf{87.0}    & \textbf{76.6}   & \textbf{87.0}   \\
Deepseek-v3             & 61.6     & 76.6    & 69.2   & 67.4   \\
Qwen2.5-7B              & 47.0     & 56.0    & 61.7   & 45.6   \\ \midrule
\textit{Fine-tuned}                                            \\ \midrule
PandaLM-7B              & 14.9     & 21.2    & 48.9   & 18.5   \\
Auto-J-13B              & 20.5     & 21.2    & 47.9   & 21.7   \\
Prometheus2-7B          & 25.4     & 31.0    & \textbf{63.8} & 45.6   \\
JudgeLM-7B              & 21.3     & 25.5    & 41.5   & 23.9   \\
Themis-8B               & 20.2     & 32.6    & 31.9   & 21.7   \\
ARJudge                & \textbf{72.4}& \textbf{73.4} & 60.7   & \textbf{67.4} \\ \bottomrule
\end{tabular}
}
\caption{Evaluation accuracy on test subsets of LLMBar series. The highest average accuracy is marked by \textbf{bold}.}
\label{llmbar_result}
\end{table}

\subsection{Implementation Details}
To construct the Composite Analysis Corpus, we prompt GPT-4o to generate evaluation questions for each instruction and collect text-based analysis. Besides, we prompt Claude-3.5-Sonnet to generate Python functions for code-driven objective analysis. We selected Claude-3.5-Sonnet due to its superior performance in code generation. 
We fine-tune Qwen2.5-7B-Instruct \cite{qwen25} on the corpus, creating a model we refer to as the Analyzer for performing multi-faceted evaluations. We use the same model in a zero-shot setting as the Refiner, with carefully crafted prompt templates. All generation in the main experiments is performed using greedy decoding by setting the temperature to 0. Details are described in Appendix \ref{train_detail}.

\subsection{Benchmarks}

We assess our framework on various evaluation datasets. Four human-annotated pairwise evaluation test sets are included: PandaLM Eval \cite{pandalm}, Auto-J Eval \cite{auto-j}, MTBench \cite{mtbench}, and the LLMBar series \cite{llmbar}. These sets were chosen for their broad coverage of evaluation tasks and their diverse set of evaluation criteria. For the LLMBar series, we use four adversarial sets, Neighbor, GPTInst, GPTOut, and Manual, as unseen sets. Unlike the other three sets and our training datasets, where candidate responses are directly sampled based on instructions, the responses in LLMBar are artificially designed to challenge evaluators by incorporating potentially misleading qualities, such as a more engaging tone. One GPT-4-annotated pairwise evaluation set, JudgeLM Eval \cite{judgelm}, is adopted. For all pairwise sets, samples with two equally preferred responses were omitted. Additionally, an instruction-following benchmark, IFEval \cite{ifeval}, is incorporated. We use this benchmark to assess the effectiveness of code-driven analysis.

\subsection{Baselines}

\paragraph{Tuning-free General LLMs}
We compare our framework with several general LLMs that can evaluate response quality. Three powerful LLMs, GPT-4o, Deepseek-v3 \cite{deepseek-v3}, and Claude-3.5-Sonnet, are used due to their balanced and comprehensive performance across most evaluation tasks \cite{sft_eval_limit}. Additionally, the backbone model used for fine-tuning the Analyzer, Qwen2.5-7B-Instruct \cite{qwen25}, is employed to demonstrate improvements. The implementation of closed-source models is done via their respective APIs.

\paragraph{Fine-tuned Evaluators}
We employ five fine-tuned evaluation models that can conduct pairwise evaluation. PandaLM \cite{pandalm} compares two responses and identifies the better one. Auto-J \cite{auto-j} and Prometheus \cite{prometheus2} support both single-response scoring and pairwise response comparison. Themis \cite{themis} rates each response based on various criteria and determines the better one by comparing their scores. JudgeLM \cite{judgelm} provides a comparison of two responses along with their corresponding scores. We use official models with 7B parameters for PandaLM, Prometheus, and JudgeLM, and models with 13B and 8B parameters for Auto-J and Themis, respectively.

\subsection{Main Results}

The main comparative results against baseline methods are shown in Table \ref{main_result}. Following \citet{llmbar} and \citet{auto-j}, we calculate the accuracy of the pairwise preference evaluation with and without swapping the two candidate responses, respectively. The average accuracy and the positional agreement rate are displayed as \textbf{Acc} and \textbf{Agr}. The performance in LLMBar is the average of its four subsets. We observe that ARJudge surpasses all fine-tuned evaluators of similar model sizes. Notably, on the challenging LLMBar set, ARJudge outperforms the best fine-tuned baseline, Prometheus2-7B, by 26.7\%. Even without more exposure to challenging samples like LLMBar, ARJudge achieves an average 15.6\% improvement over its backbone model, Qwen2.5-7B-Instruct. Additionally, ARJudge's performance is comparable to that of powerful tuning-free LLMs on some test sets. For example, ARJudge performs on par with GPT-4o and Claude-3.5-Sonnet on Auto-J Eval and with DeepSeek-V3 on LLMBar. Besides, compared to other fine-tuned methods, ARJudge can generalize to more test sets.

Table \ref{llmbar_result} further presents detailed evaluation results in different subsets of LLMBar. Our ARJudge performs the best on most subsets and has made significant improvements compared to the backbone model, Qwen2.5-7B-Instruct. On LLMBar-Neighbor, it achieves higher evaluation accuracy than the advanced DeepSeek-V3. 

\subsection{Ablation Study}
\begin{table}[t!]
\center
\resizebox{1.0\linewidth}{!}{
\begin{tabular}{@{}lccccc@{}}
\toprule
Models         & JudgeLM  & PandaLM & Auto-J & MTBench & LLMBar \\ \midrule
Qwen2.5-7B     & 80.0  & 80.7    & 73.8   & 75.2    & 52.6   \\ \midrule
ARJudge        & 81.0 & 82.4    & 78.5   & 78.3    & 68.2   \\
-w/o FT        & 73.1 & 75.6    & 68.7   & 70.0    & 62.5   \\
-w/o FT\&MF    & 74.7 & 72.2    & 65.6   & 67.8    & 63.7   \\
-w/o Refine    & 81.7 & 82.8    & 79.6   & 79.1    & 63.7   \\ \bottomrule
\end{tabular}
}
\caption{Comparison results under ablation settings. ``JudgeLM'', ``PandaLM'', and ``Auto-J'' are abbreviation of the associated testsets. ``\textbf{FT}'' and ``\textbf{MF}'' represent fine-tuning and multi-faceted.}
\label{ablation}
\end{table}

To further investigate the effectiveness of our framework, we analyze several variations of ARJudge, as detailed below. 
(1) \textbf{w/o FT}: we replace the fine-tuned Analyzer with the same tuning-free model as the Refiner and prompt the model to generate evaluation questions and conduct the multi-faceted evaluation. 
(2) \textbf{w/o FT\&MF}: we apply the model as in the w/o FT setting, generating Chain-of-Thought (CoT) evaluations directly. 
(3) \textbf{w/o Refine}: we retain the fine-tuned Analyzer and make slight modifications to the prompt for the Refiner to directly output the label of the better response. 

The ablation results are shown in Table \ref{ablation}. We observe accuracy drops across all test sets with the ablation variants, indicating the effectiveness of each component in ARJudge. Specifically, fine-tuning significantly enhances a general LLM’s evaluation capability, enabling it to propose reasonable evaluation questions and analyze responses accordingly. Evaluation questions help the LLM focus on relevant aspects and enhance its evaluation performance. Interestingly, we find that the effects of refinement differ between the fine-tuned and tuning-free Analyzer. In JudgeLM Eval, PandaLM Eval, Auto-J, and MTBench, the refinement keeps evaluation accuracy under the fine-tuned Analyzer’s analysis (w/o Refine vs. ARJudge) but significantly decreases it under the tuning-free Analyzer’s analysis (Qwen2.5-7B vs. w/o FT\&MF). It may be related to the controversial phenomenon that LLMs cannot truly self-correct \cite{self-correct}. Additionally, for challenging samples in LLMBar, refinement significantly strengthens the performance of the fine-tuned and tuning-free ones.

\subsection{Capability to Evaluate Using Code}

\begin{figure}[t!]
	\centering
	\includegraphics[width=1\linewidth]{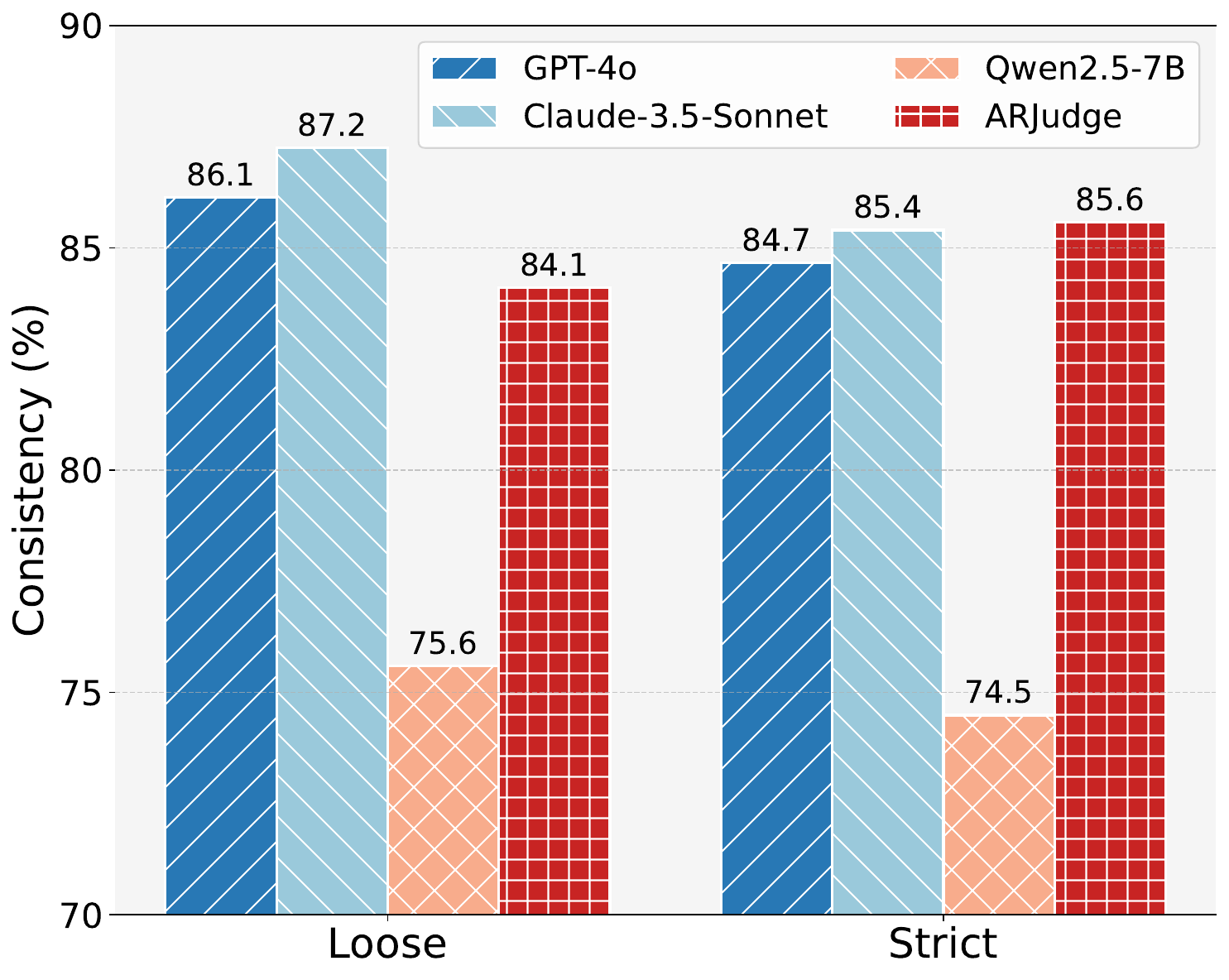}
	\caption{Results on the consistency between code-driven evaluation and IFEval evaluation. ``Loose'' and ``Strict'' are two judgment criteria in IFEval.}
	\label{code_eval}
\end{figure}

Code-driven analysis plays a crucial role in robustly verifying the objective requirements of instructions. To assess the effectiveness of code-driven analysis, we use the execution results of the IFEval official code as a benchmark and compute the \textbf{Consistency} between its judgment (Loose or Strict) and that of other models. We compare ARJudge with GPT-4o, Claude-3.5-Sonnet, and Qwen2.5-7B-Instruct. These three models are prompted to make judgments in a zero-shot manner. As shown in Figure \ref{code_eval}, our framework achieves a significant improvement over the backbone model, Qwen2.5-7B-Instruct, with the help of code-driven analysis. Moreover, ARJudge performs comparably to GPT-4o and Claude-3.5-Sonnet, demonstrating its potential as a viable alternative. Notably, the execution success rate of the generated code is 100\%. 

\subsection{Effect of Increasing Analysis Quantity}\label{effect_q}

\begin{figure}[t!]
	\centering
	\includegraphics[width=1.0\linewidth]{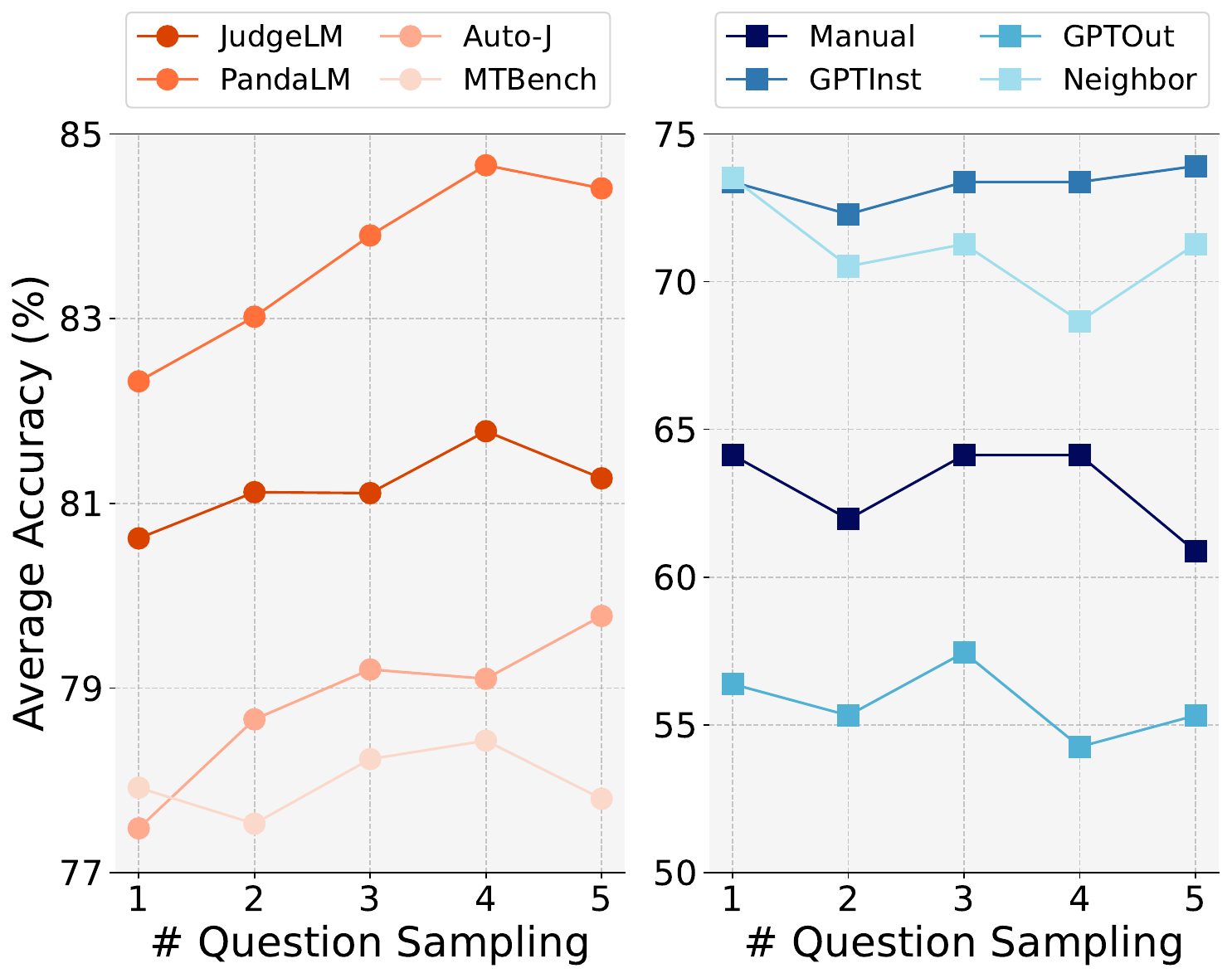}
	\caption{Evaluation results with increasing analyses. The right displays the results of four subsets in LLMBar.}
	\label{num_questions}
\end{figure}

We extend our analysis by scaling up the number of question sampling attempts, exploring the effect of increasing analysis quantity. We set the temperature to 0.2 to sample evaluation questions multiple times, ensuring diversity in the generated questions. As shown in Figure \ref{num_questions}, evaluation accuracy improves with more analyses for most datasets, including JudgeLM Eval, Auto-J Eval, PandaLM Eval, and MTBench. The highest accuracy is achieved with four or five rounds of question sampling and their combined analysis. However, in the LLMBar series, additional analysis had little or even a negative impact on accuracy. This may be because the Analyzer has greater uncertainty about the evaluation samples in these sets, and additional analysis further amplifies this uncertainty.

\subsection{Generalization of Evaluation Capability}\label{effect_refine}

To further demonstrate the generalization of evaluation capability, we compute the ratio of judgment change after refining as shown in Table \ref{refine}. Combining Table \ref{ablation} and \ref{refine}, we observe that the Refiner maintains evaluation performance in JudgeLM Eval, PandaLM Eval, Auto-J Eval, and MTBench, while significantly increasing it in the LLMBar series. This indicates that re-analysis improves the generalization of evaluation capability, especially in handling unseen challenging samples.

\section{Case Studies}

\begin{table}[t!]
\center
\resizebox{1.0\linewidth}{!}{
\begin{tabular}{@{}lccccc@{}}
\toprule
Models           & JudgeLM  & PandaLM & Auto-J & MTBench & LLMBar \\ \midrule
W$\rightarrow$C  & 3.9 & 4.4    & 2.3   & 2.1    & 7.8   \\
C$\rightarrow$W  & 4.6 & 4.8    & 3.4   & 2.9    & 3.6   \\
\bottomrule
\end{tabular}
}
\caption{Ratio of change after refining. ``W$\rightarrow$C'' denotes a judgment changing from wrong to correct after refinement, while ``C$\rightarrow$W'' denotes the opposite.}
\label{refine}
\end{table}

\begin{figure*}[t!]
	\centering
	\includegraphics[width=1.0\linewidth]{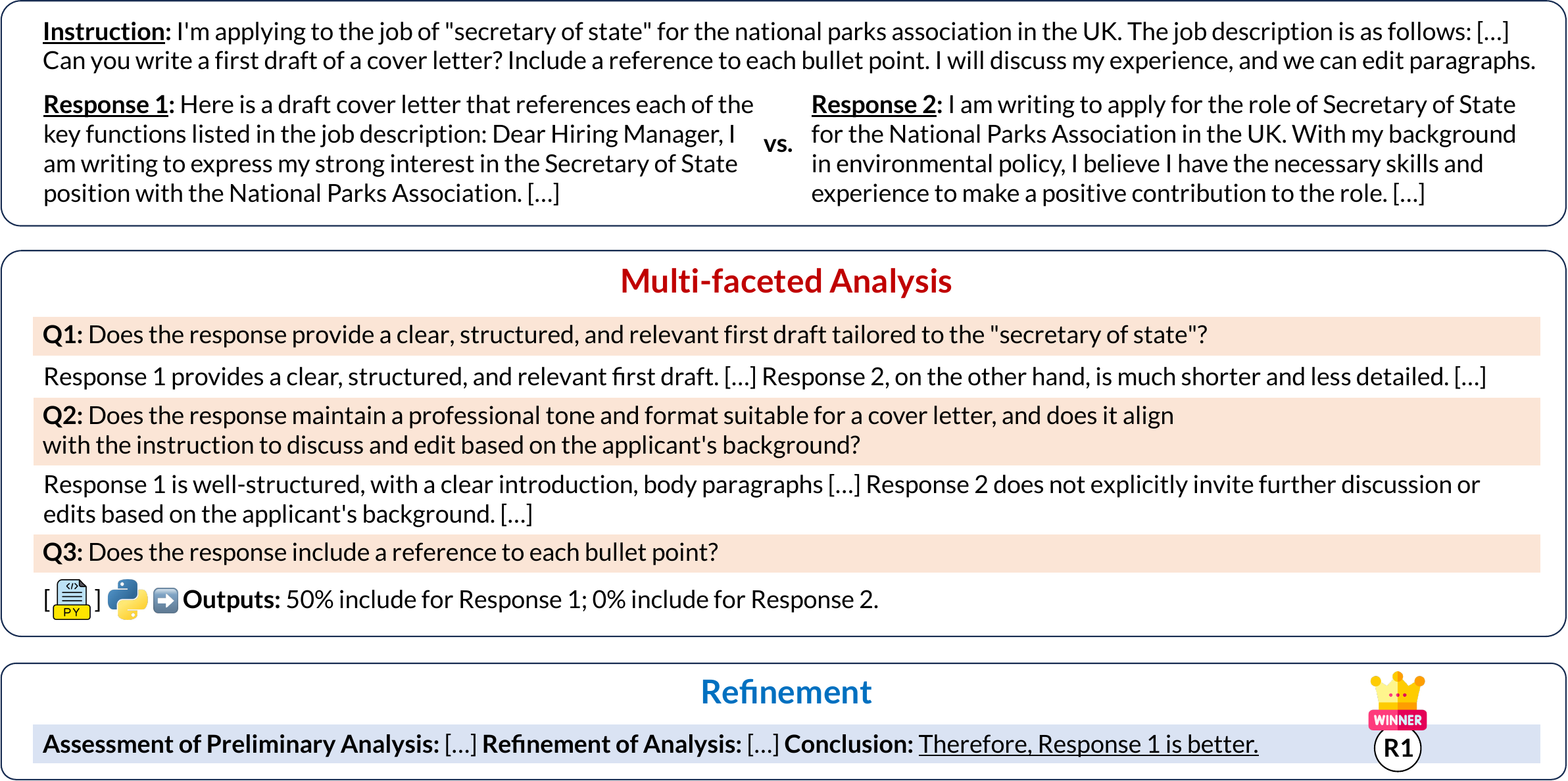}
	\caption{An example of evaluation generated by ARJudge.}
	\label{case_study}
\end{figure*}

We show an example of a multi-faceted evaluation generated by ARJudge in Figure \ref{case_study}. Given an instruction and two responses, the Analyzer first generates three evaluation questions and the corresponding multi-faceted analyses. The last question is analyzed by constructing a Python function and assessing execution feedback to determine requirement completeness. Then, the Refiner reviews the preliminary analysis and refines it by reconsidering the instruction's requirements. 
\section{Related Work}
\subsection{Tuning-Free Generalist Evaluators}
Tuning-free generalist evaluators leverage the inherent capabilities of large language models (LLMs) to assess responses through the use of carefully designed prompts, offering exceptional flexibility and scalability. Various techniques have been employed to enhance the accuracy of these evaluations, such as in-context learning \cite{Fu2023GPTScoreEA,lin2023llm}), adding task-specific criteria~\cite{kotonya2023little,zhuo2024ice}, and Chain-of-Thought analysis~\cite{liu2023g,zhuo2024ice}). 

Despite their versatility, tuning-free evaluators often suffer from biases such as position bias~\cite{raina2024llm,wang2023large,zheng2023judging} and verbosity bias~\cite{khandebating,ye2024justice}, which can skew evaluation outcomes. Methods like response-adapted references~\cite{zhang2024reviseval}, multi-agent collaboration~\cite{xu2023towards}, and divide and conquer~\cite{branch-merge,li2023split} have been proposed to mitigate these issues, improving the fairness and reliability of LLM-based evaluations.

\subsection{Specialized Fine-Tuned Evaluators}
While tuning-free approaches provide flexibility, specialized fine-tuned evaluators are explicitly trained on human-labeled preference data to achieve higher accuracy and domain-specific reliability. These models undergo supervised fine-tuning or reinforcement learning-based optimization to align their evaluations more closely with expert judgments~\cite{auto-j,pandalm,prometheus,prometheus2, sorry-bench}.

While fine-tuned evaluators offer improved accuracy, they face notable challenges in scalability and generalization~\cite{sft_eval_limit}. Unlike tuning-free approaches, which can adapt to new tasks with minimal configuration, fine-tuned models require ongoing updates through methods such as supervised fine-tuning or direct preference optimization~\cite{rafailov2024direct}. To remain effective amidst evolving benchmarks~\cite{mtbench,llmbar}, Auto-J~\cite{auto-j} leverages a large dataset of scoring and preference annotations while incorporating dynamic in-context learning techniques, such as few-shot prompting, to enhance adaptability. Similarly, FLAMe~\cite{vu2024foundational} combines fine-tuning on labeled preference data with large-scale multitask instruction tuning, enabling it to dynamically adapt to new evaluation criteria while maintaining flexibility.

\section{Conclusion}

This work proposes a novel evaluation framework, ARJudge, which adaptively designs evaluation criteria and performs multi-faceted evaluation in both text and code. A new Composite Analysis Corpus, designed for both criteria generation and multi-faceted analysis, is developed to train ARJudge. Extensive experiments demonstrate the superiority and robustness of our framework across diverse evaluation benchmarks. Notably, with code-driven analyses, ARJudge gains strong evaluation capabilities for assessing instruction following. Future studies can explore the effective use of more tools, such as a search engine, to improve evaluation honesty and mitigate hallucination.

\section*{Limitations}

While our framework outperforms various baseline approaches in LLM evaluation, there is still room for improvement. Our method is limited to using code to enhance evaluation robustness and does not consider additional tools such as search engines or specialized agents. Furthermore, our approach partially relies on the LLM's own reasoning ability for evaluation. If the LLM itself lacks strong reasoning capabilities, the effectiveness of refinement may be limited. Additionally, our evaluation is restricted to pairwise comparisons and does not enhance the model's ability to score single responses. Although single-response scoring can be achieved by modifying the Refiner’s prompt, its accuracy has not been properly aligned.

\section*{Acknowledgment}
This work was supported by the Research Grants Council of Hong Kong (15207920, 15213323). 

\bibliography{custom}

\appendix
\section{Training Settings}\label{train_detail}

We train Qwen2.5-7B-Instruct\footnote{https://huggingface.co/Qwen/Qwen2.5-7B-Instruct} to perform as the Analyzer. The number of training samples in the Composite Analysis Corpus is around 25K, including 7.7K evaluation question generation samples, 6K code-driven analysis samples, and 11K text-based analysis samples. The corpus is constructed based on instructions from Auto-J\footnote{https://github.com/GAIR-NLP/auto-j} \cite{auto-j}. We train it for 2 epochs with a global batch size of 96 and we save checkpoints for every 50 steps. The learning rate is set to 1e-5. We use DeepSpeed ZeRO3 and FlashAttention to reduce computational memory usage. The training is implemented on 6 computing devices. We use Pytorch with the 2.4.0 version, Transformers with the 4.44.2 version, and deepspeed with the 0.14.4 version. 

\section{Prompt Templates}

\begin{figure*}[t!]
	\centering
	\includegraphics[width=1.0\linewidth]{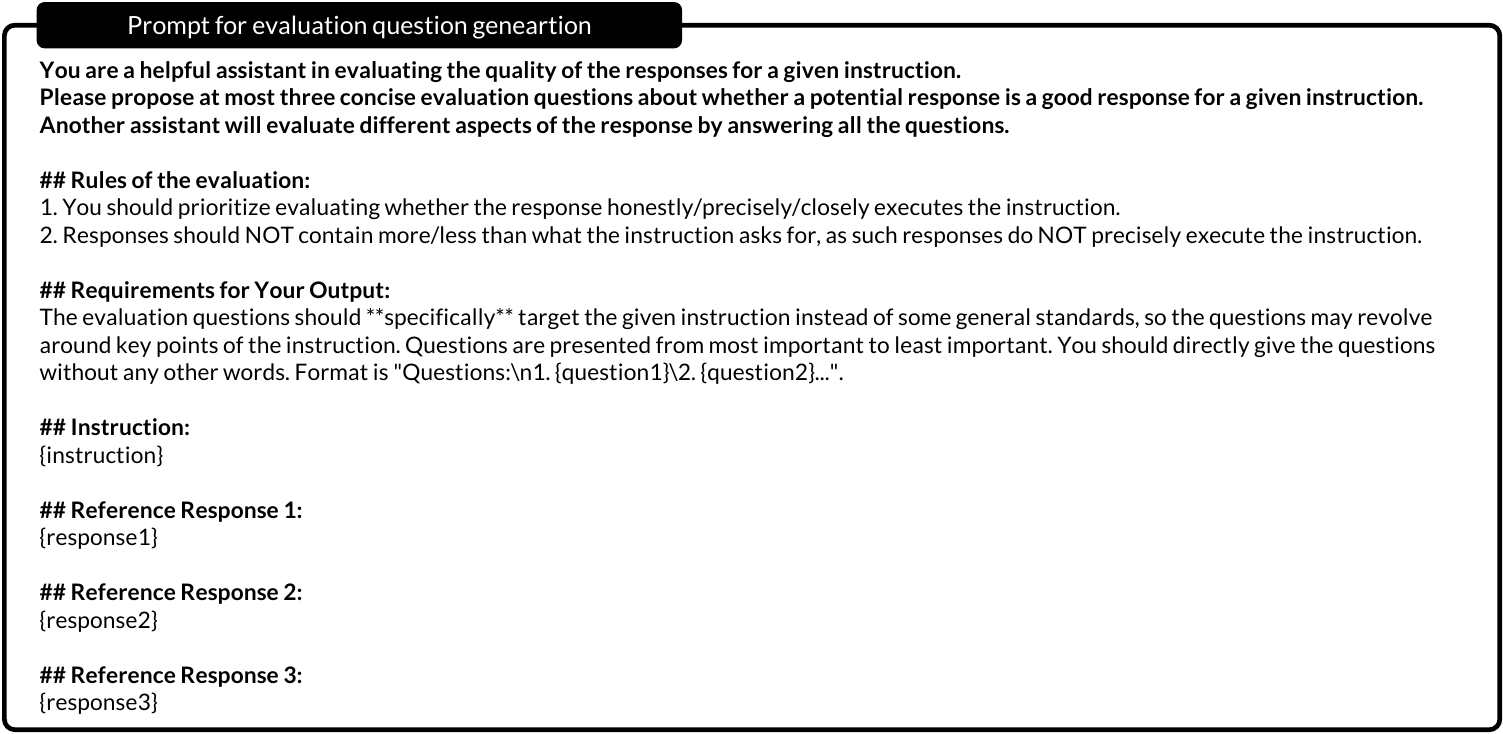}
	\caption{Prompt template for evaluation question generation.}
	\label{prompt_question}
\end{figure*}

\begin{figure*}[t!]
	\centering
	\includegraphics[width=1.0\linewidth]{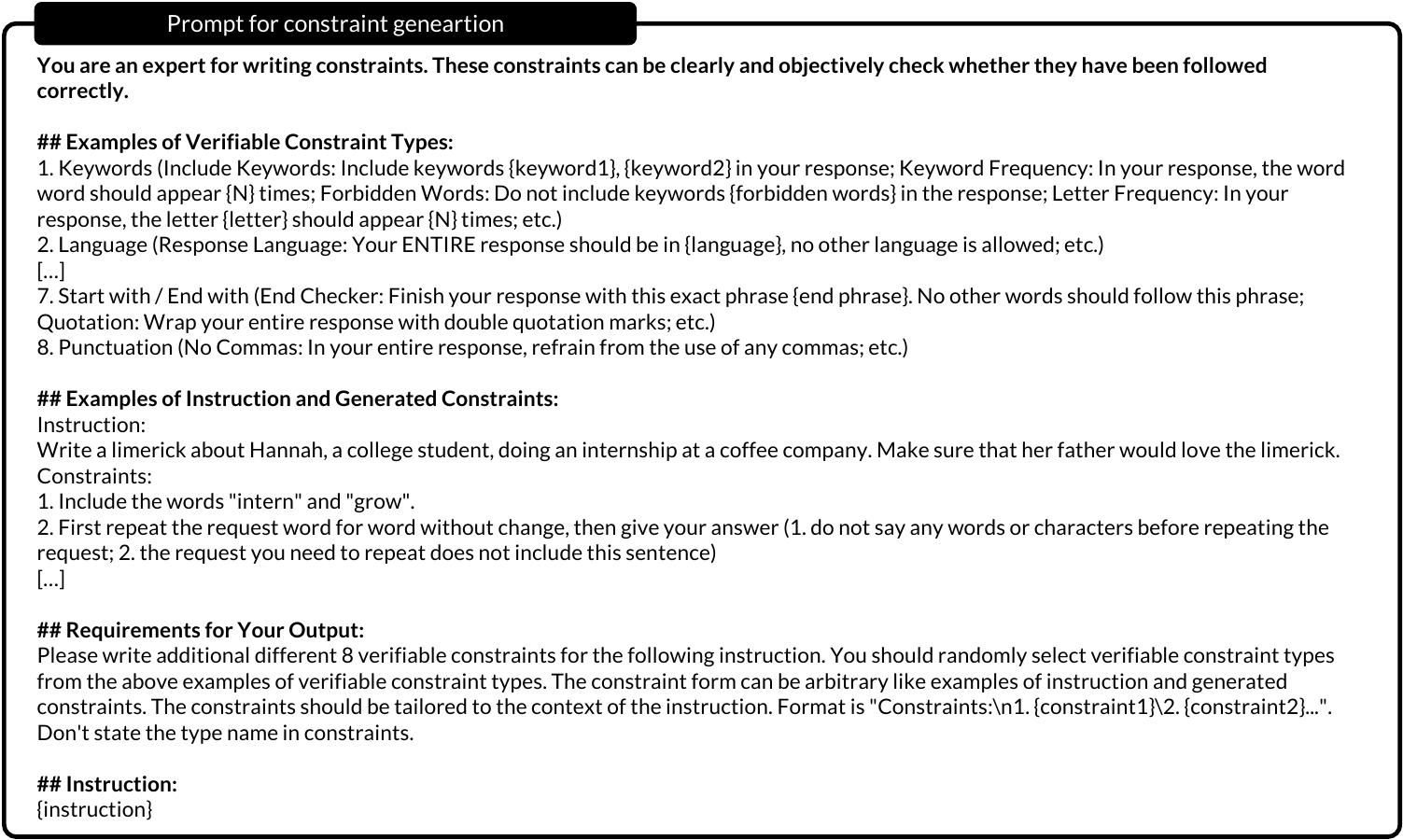}
	\caption{Prompt template for objective constraint generation.}
	\label{prompt_constraint}
\end{figure*}

\begin{figure*}[t!]
	\centering
	\includegraphics[width=1.0\linewidth]{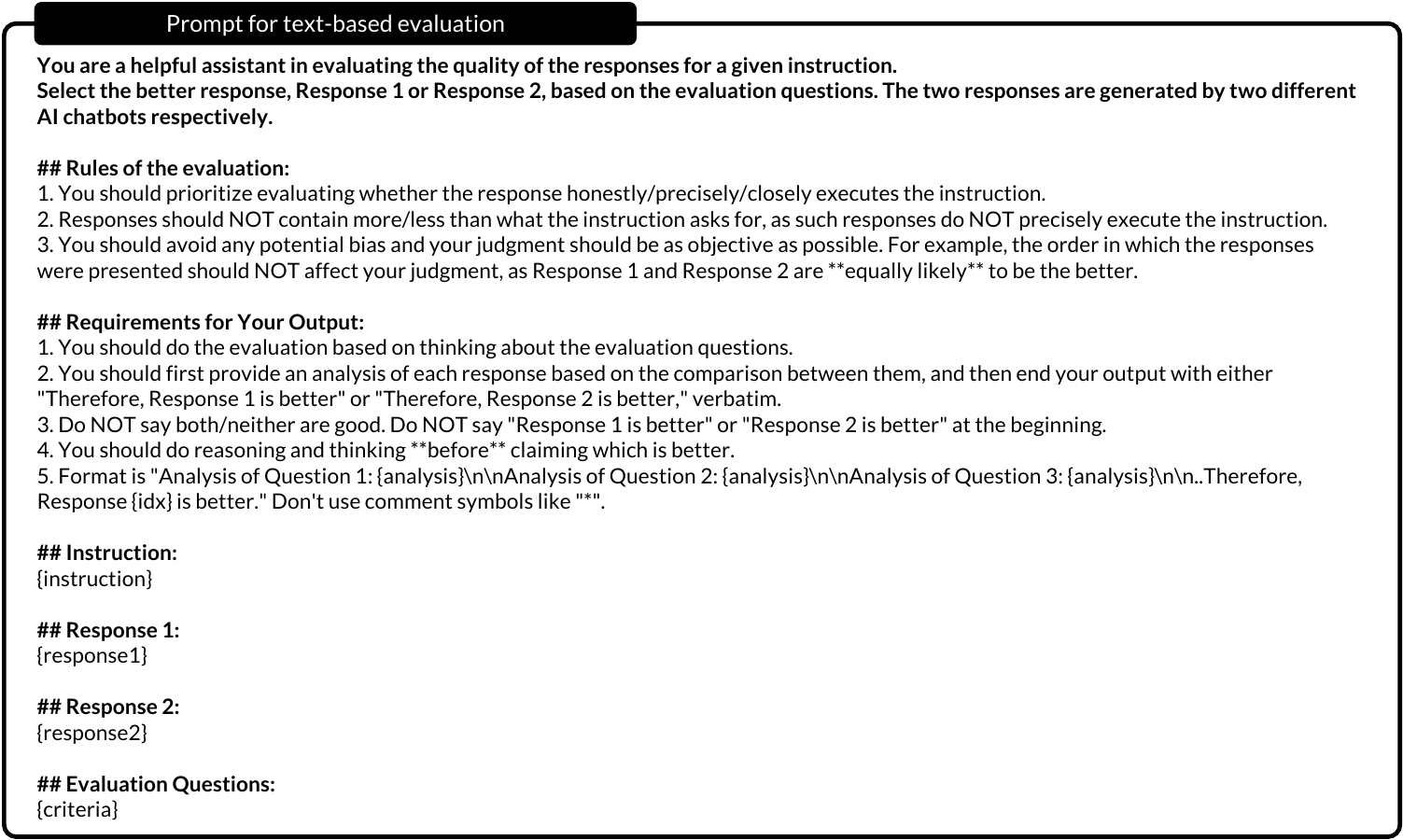}
	\caption{Prompt template for text-based evaluation.}
	\label{prompt_text_eval}
\end{figure*}

\begin{figure*}[t!]
	\centering
	\includegraphics[width=1.0\linewidth]{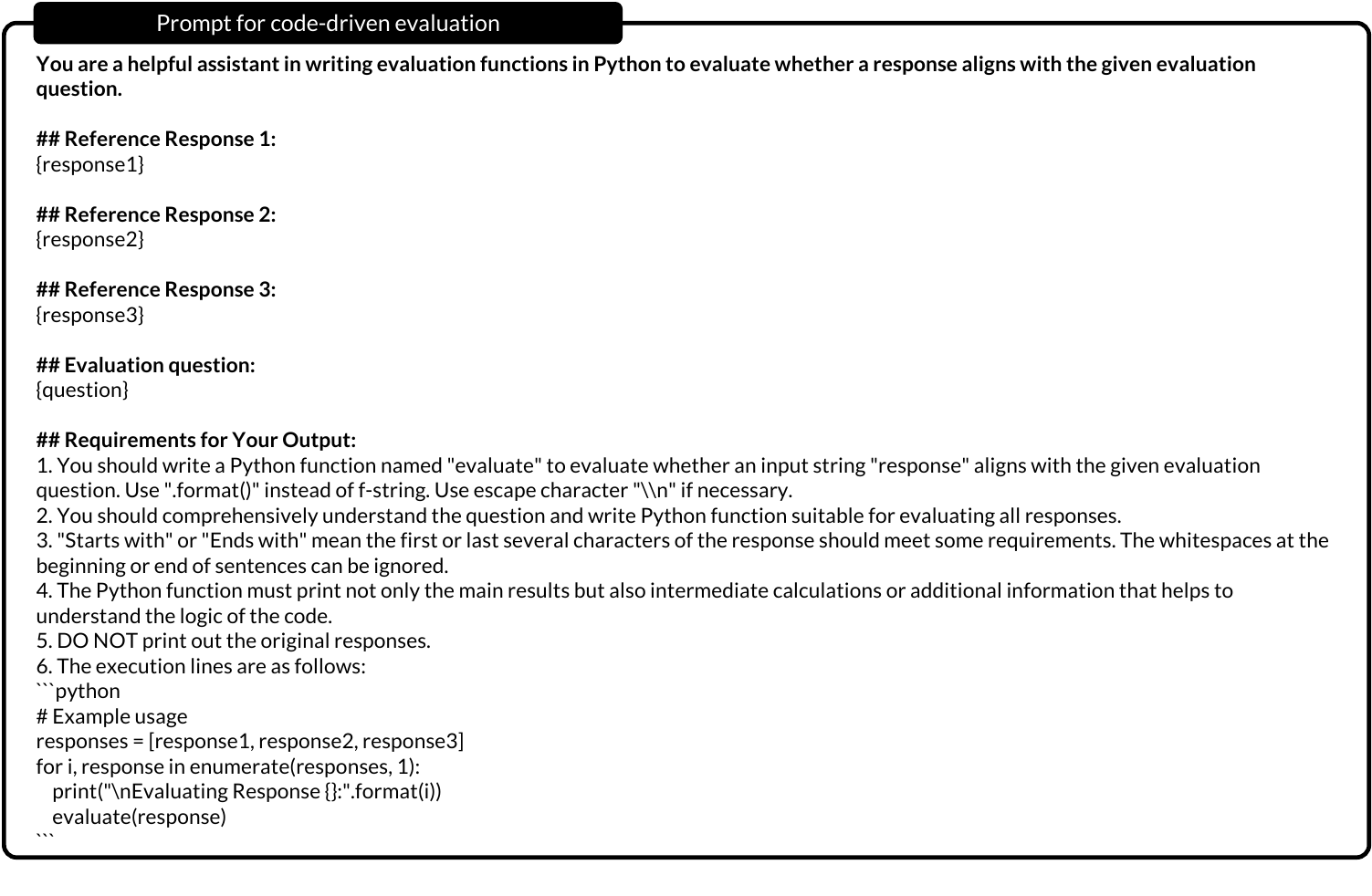}
	\caption{Prompt template for code-driven evaluation.}
	\label{prompt_code_eval}
\end{figure*}

\begin{figure*}[t!]
	\centering
	\includegraphics[width=1.0\linewidth]{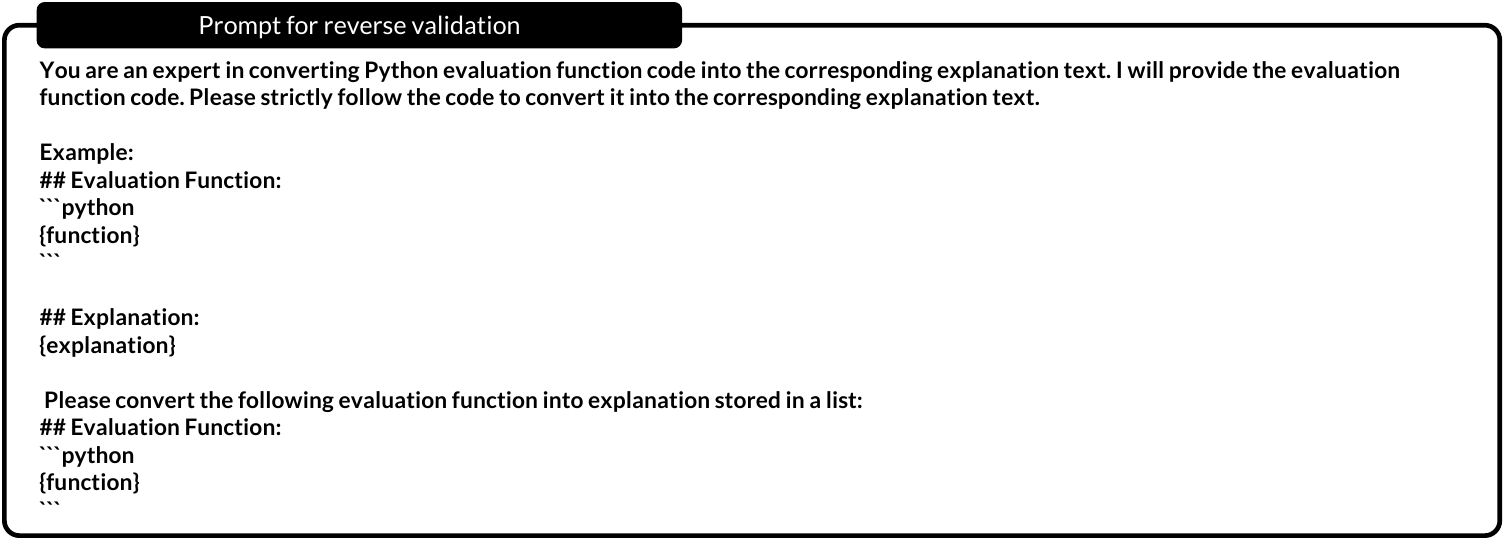}
	\caption{Prompt template for reverse validation.}
	\label{prompt_reverse}
\end{figure*}

Prompt templates used for dataset construction are shown in Figure \ref{prompt_question}, Figure \ref{prompt_constraint}, Figure \ref{prompt_text_eval}, Figure \ref{prompt_code_eval}, and Figure \ref{prompt_reverse}.

\begin{figure*}[t!]
	\centering
	\includegraphics[width=1.0\linewidth]{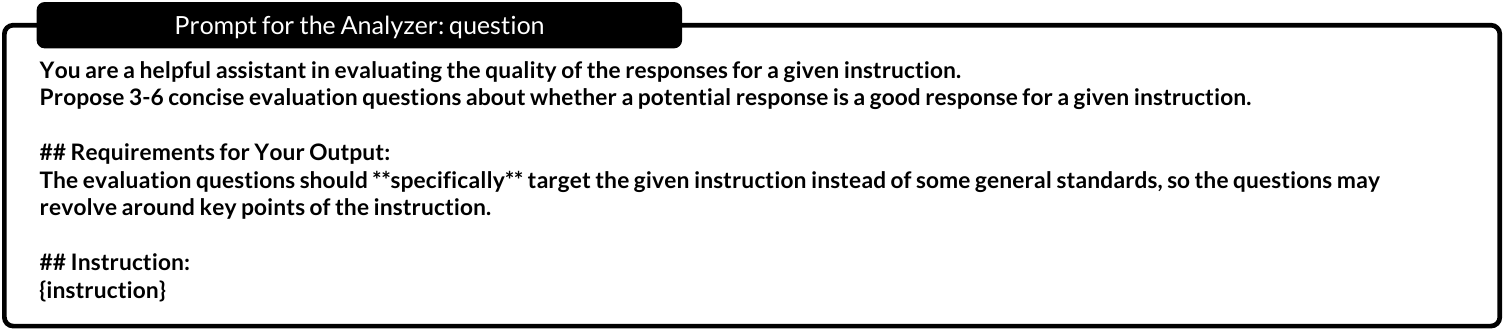}
	\caption{Prompt template for question generation of the Analyzer.}
	\label{prompt_analyzer_question}
\end{figure*}

\begin{figure*}[t!]
	\centering
	\includegraphics[width=1.0\linewidth]{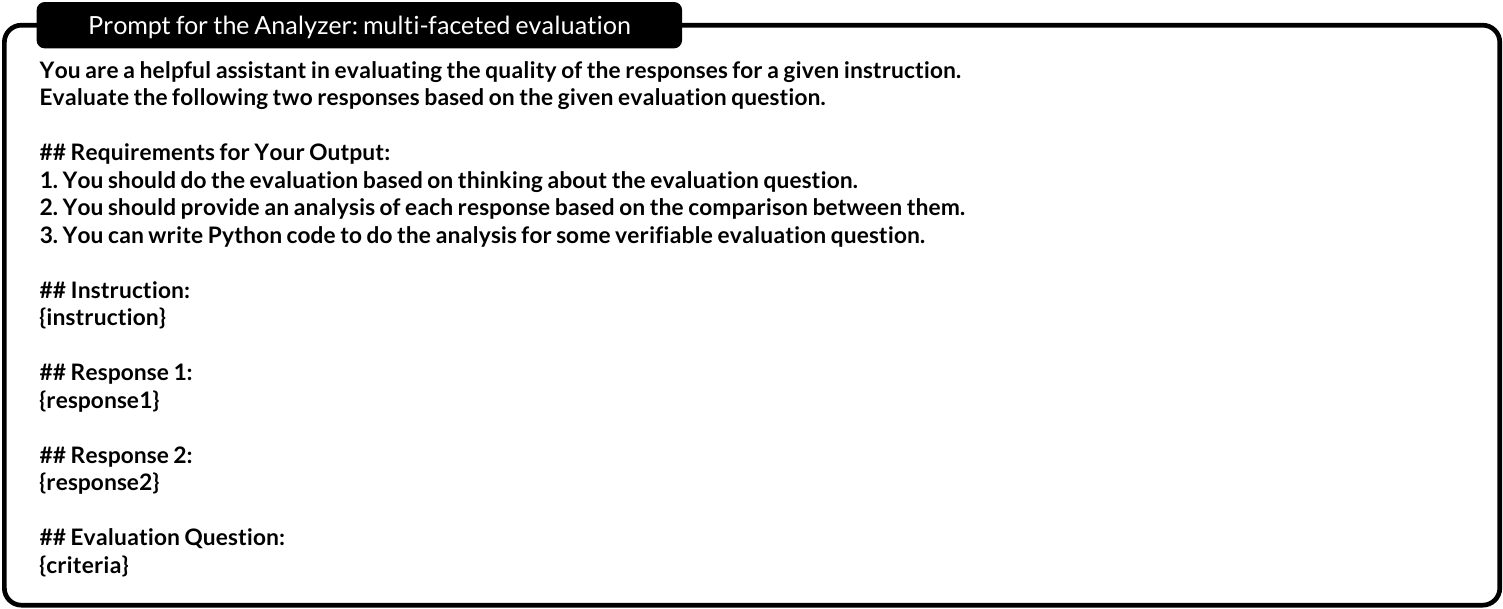}
	\caption{Prompt template for multi-faceted analysis of the Analyzer.}
	\label{prompt_analyzer_analysis}
\end{figure*}

\begin{figure*}[t!]
	\centering
	\includegraphics[width=1.0\linewidth]{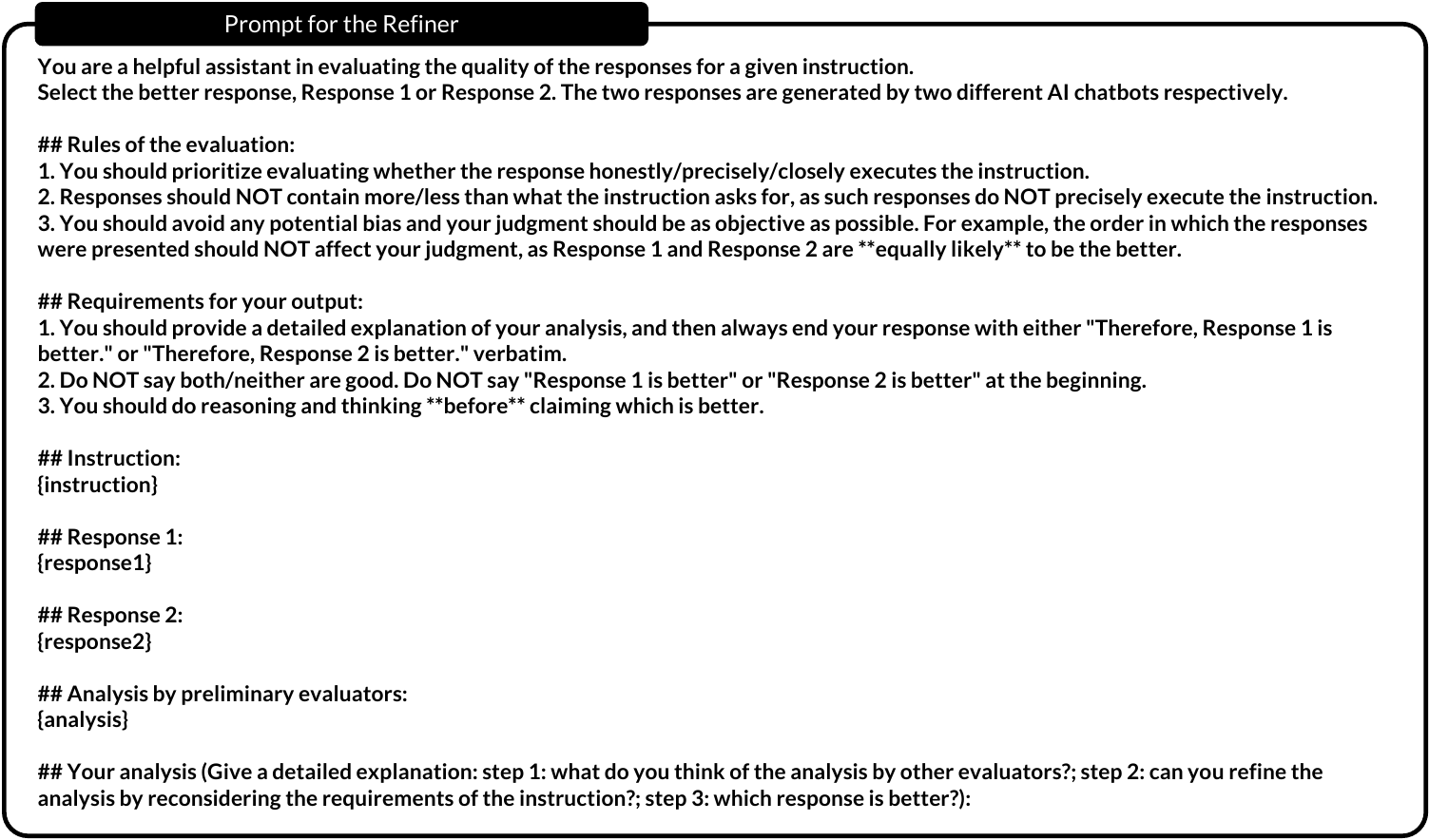}
	\caption{Prompt template for refinement of the Refiner.}
	\label{prompt_refiner}
\end{figure*}

Prompt templates used for the Analyzer and Refiner of our ARJudge are shown in Figure \ref{prompt_analyzer_question}, Figure \ref{prompt_analyzer_analysis}, and Figure \ref{prompt_refiner}.

\section{Quality Analysis of Data Generation}

Table \ref{filtering_rates} displays the success rate of each step in the filtering process.

\begin{table}[h!]
\centering
\resizebox{1.0\linewidth}{!}{
\begin{tabular}{lc}
\toprule
\textbf{Filtering} & \textbf{Remaining rate} \\ \midrule
Check 1 (execute with the sample responses in the prompt)   & 100\% \\
Check 2 (execute with another three sample responses)       & 99\%  \\
Reverse Validation                                          & 94\%  \\ \bottomrule
\end{tabular}
}
\caption{Filtering Process and Corresponding Remaining Rates}
\label{filtering_rates}
\end{table}

Overall, the high remaining rates across all steps indicate the effectiveness of our generation pipeline and the reliability of the final dataset. The reverse validation step, in particular, plays a crucial role in maintaining high-quality standards.

To enhance reproducibility, we have included representative examples of both accepted and rejected Python scripts in the supplementary material, along with the rationale for their inclusion or exclusion. 

\section{Training Dataset Statistics}

Table \ref{training_stat} presents a comprehensive summary of dataset statistics, including the distribution of task types, average length of analysis, and diversity of criteria. This provides clearer insight into the scope and representativeness of the training data.

\begin{table}[h!]
\centering
\resizebox{1.0\linewidth}{!}{
\begin{tabular}{lcc}
\toprule
\textbf{Training Samples} & \textbf{Quantity} & \textbf{Average Length} \\ \midrule
Total                                & 35144 & 111.5 \\
- Text Analysis Samples              & 21060 & 81.7  \\
- Python Script Samples              & 6322  & 247.9 \\
- Evaluation Question Samples        & 7762  & 81.2  \\ \bottomrule
\end{tabular}
}
\caption{Statistics of Training Samples by Category}
\label{training_stat}
\end{table}

\section{More Statistics of Type 2 Questions}

Table \ref{constraint_types} provides detailed statistics on Type 2 question coverage. Specifically, we categorize objective constraints into 8 types following \cite{ifeval} and report their proportions in the fine-tuning set. This addition helps illustrate how ARJudge generalizes across diverse constraint-based evaluations. 

\begin{table}[h!]
\centering
\resizebox{1.0\linewidth}{!}{
\begin{tabular}{lrr}
\toprule
\textbf{Constraint Type} & \textbf{Count} & \textbf{Percentage} \\ \midrule
Keywords                        & 1,983 & 26.2\% \\
Language                        & 303   & 4.0\%  \\
Length Constraints              & 1,193 & 15.8\% \\
Detectable Content              & 627   & 8.3\%  \\
Detectable Format               & 1,035 & 13.7\% \\
Change Cases                    & 374   & 4.9\%  \\
Start with / End with          & 1,741 & 23.0\% \\
Punctuation                     & 316   & 4.2\%  \\ \midrule
\textbf{Total}                 & \textbf{7572} & \textbf{100\%} \\ \bottomrule
\end{tabular}
}
\caption{Constraint Types with Corresponding Counts and Percentages}
\label{constraint_types}
\end{table}

\section{Further Explanation of Positional Agreement Rate}

While ARJudge shows a slightly lower Agr compared to Auto-J-13B, it is important to note that Auto-J-13B is nearly twice the size of our model and uses samples with swap responses to enhance positional agreement. When compared to models of similar scale—such as PandaLM-7B, Prometheus2-7B, and JudgeLM-7B—ARJudge achieves superior or comparable Average Agr scores as shown in Table \ref{average_agreement}.

\begin{table}[h!]
\centering
\resizebox{1.0\linewidth}{!}{
\begin{tabular}{lc}
\toprule
\textbf{Model} & \textbf{Average Agr} \\ \midrule
GPT-4o             & 85.38 \\
Claude-3.5-Sonnet  & 88.54 \\
Deepseek-v3        & 85.30 \\
Qwen2.5-7B         & 72.02 \\
PandaLM-7B         & 76.65 \\
Auto-J-13B         & 85.48 \\
Prometheus2-7B     & 78.65 \\
JudgeLM-7B         & 81.33 \\
Themis-8B          & --    \\
ARJudge            & 80.85 \\ \bottomrule
\end{tabular}
}
\caption{Average Agreement Scores Across Different Models}
\label{average_agreement}
\end{table}

We believe the current level of positional agreement already demonstrates strong alignment. More importantly, ARJudge consistently outperforms all baselines across every evaluation benchmark, particularly on the more challenging LLMBar dataset (ARJudge Acc: 68.2, others < 42). This highlights the effectiveness of our multi-faceted analysis and refinement design, which contributes to more robust and accurate evaluations.

\section{Granular Ablation of Analysis Types}

Table \ref{types_analysis} presents a more fine-grained ablation study that evaluates a model variant with code-driven analysis disabled, in order to isolate the contribution of each analysis type. We do not evaluate a setting with only code-driven analysis, as most existing benchmarks are primarily designed for assessing content quality rather than objective constraints.

\begin{table}[h!]
\centering
\resizebox{1.0\linewidth}{!}{
\begin{tabular}{lccc}
\toprule
\textbf{ARJudge} & \textbf{MTBench (Acc)} & \textbf{Auto-J (Acc)} & \textbf{IFEval (Consistency)} \\ \midrule
Complete   & 78.3 & 78.5 & 85.6 \\
Text Only  & 77.4 & 77.9 & 76.1 \\ \bottomrule
\end{tabular}
}
\caption{Performance Comparison of ARJudge Methods across Different Benchmarks}
\label{types_analysis}
\end{table}

The results demonstrate that code-driven analysis contributes meaningfully to performance, particularly on MTBench, Auto-J Eval, and IFEval. The variation in performance gains across these benchmarks reflects their differing focuses: MTBench and Auto-J Eval emphasize open-ended content evaluation, where textual coherence and reasoning dominate, while IFEval targets instruction-following with verifiable constraints, where code-based analysis plays a more central role.

\end{document}